\documentclass[11pt,a4paper]{article}
\usepackage{acl}
\usepackage{times}
\usepackage{latexsym}

\usepackage{slashbox}
\usepackage{multirow}
\usepackage{hhline}
\usepackage{subcaption}
\usepackage{microtype}
\usepackage{float}
\usepackage{url}
\usepackage{booktabs}
\usepackage{amsfonts}
\usepackage{amsmath,amsthm}
\usepackage{graphicx}
\usepackage{mathrsfs}
\usepackage{enumitem}

\usepackage{soul}
\DeclareMathOperator*{\argmax}{arg\,max}
\newcommand{\makenonemptybox}[2]{%
\fbox{
\parbox[c][#1][t]{\dimexpr\linewidth+\fboxsep+\fboxrule}{
  \hrule width \hsize height 0pt
  #2
 }%
}%
\par\vspace{\ht\strutbox}
}

\newcommand{\shaycomment}[1]{\textcolor{blue}{#1 -- Shay}}
\newcommand{\yftahcomment}[1]{\textcolor{red}{#1}}
\newcommand{\wayloncomment}[1]{\textcolor{teal}{#1}}

\newcommand{\ignore}[1]{}

\newenvironment{itemizesquish}[2]{\begin{list}{\labelitemi}{\setlength{\itemsep}{#1}\setlength{\labelwidth}{#2}\setlength{\leftmargin}{\labelwidth}\addtolength{\leftmargin}{\labelsep}}}{\end{list}}

\title{BERT is not The Count: Learning to Match \\ Mathematical Statements with Proofs}

\author{Weixian Waylon Li$^1$ $\qquad$ Yftah Ziser$^1$ $\qquad$ Maximin Coavoux$^2$\Thanks{ Work mostly done at the University of Edinburgh.} $\qquad$ Shay B. Cohen$^1$\\
$^1$ University of Edinburgh, School of Informatics, Edinburgh\\
  $^2$ Univ. Grenoble Alpes, CNRS, Grenoble INP, LIG, 
   \\
\medskip
\texttt{W.Li-67@sms.ed.ac.uk} $\qquad$ \texttt{yftah.ziser@ed.ac.uk}
}

\date{}

\begin{document}
\maketitle
\begin{abstract}
We introduce a task consisting in matching a proof to a given mathematical statement. The task fits well within current research on Mathematical Information Retrieval and, more generally, mathematical article analysis \cite{2014arXiv1404.1905M}. We present a dataset for the task (the \textsc{MATcH} dataset) consisting of over 180k statement-proof pairs extracted from modern mathematical research articles.\footnote{Our dataset and code are available at \url{https://github.com/waylonli/MATcH}.} We find this dataset highly representative of our task, as it consists of relatively new findings useful to mathematicians. We propose a bilinear similarity model and two decoding methods to match statements to proofs effectively. While the first decoding method matches a proof to a statement without being aware of other statements or proofs, the second method treats the task as a global matching problem. Through a symbol replacement procedure, we analyze the ``insights"  that pre-trained language models have in such mathematical article analysis and show that while these models perform well on this task with the best performing mean reciprocal rank of 73.7, they follow a relatively shallow symbolic analysis and matching to achieve that performance.\footnote{Like Bert, The Count (or Count von Count; \raisebox{-0.05in}{\includegraphics[width=0.21in]{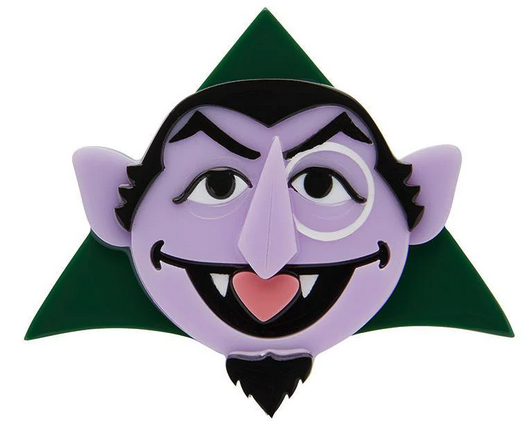}}) is a character from the television show Sesame Street. The Count likes counting, and his main role in the show is to teach this skill to children.}

\end{abstract}

\section{Introduction}

Research-level mathematical discourse is a challenging domain for
Natural Language Processing (NLP).
Mathematical articles frequently switch between natural
language and mathematical formulae, and
a semantic analysis of mathematical text needs to solve
relationships (e.g.\ coreference) between mathematical symbols
and concepts.
Moreover, mathematical writing follows many conventions,
such as variable naming or typography that are implicit,
and may differ between subfields.

However, mathematical research can benefit from NLP \cite{2014arXiv1404.1905M},
in particular as concerns bibliographical research:
researchers need tools to find work relevant to their research.
Indeed, prior NLP work on mathematical research articles
focused on Mathematical Information Retrieval (MIR)
and related tools or data \cite{DBLP:conf/ntcir/ZanibbiAKOTD16,C16-1221,P15-2055}.

\begin{figure}
  \fbox{
  \parbox{0.94\columnwidth}{
  \textbf{Statement. }
When $m=0$ we have $E^{0}_{rg}=\emptyset$, 
and when $m\neq 0$ we have $E^{0}_{rg}=E^0$.

\textbf{Proof. }
When $m=0$, the image of $r$ is $\{1\}$. 
Hence $E^{0}_{rg}=\emptyset$. 
When $m\neq 0$, the map $r$ is a surjective proper map. 
Hence $E^{0}_{rg}=E^0$. 


  }
}
  \caption{Example of a statement-proof pair.}
  \label{fig:pair-example}
\end{figure}

We introduce a task
aimed at improving the processing of research-level
mathematical articles and make a step towards the
modeling of mathematical reasoning.
Given a collection of mathematical statements
and a collection of mathematical proofs of the same
size, the task consists in finding and assigning
a proof to each mathematical statement.
We construct and release a dataset for the task (\textsc{MATcH}), by collecting over 180k statement-proof pairs from mathematical research articles
(an example is given in Figure~\ref{fig:pair-example}).

Related datasets, such as \textsc{LeanStep} \cite{leanstep} and the synthetic dataset of \newcite{synthetic_dataset} do not include natural language. 
NaturalProofs \cite{naturalproof}, another related dataset, only consists of 32k theorem-proof pairs from ProofWiki,\footnote{\url{https://proofwiki.org/xmldump/latest.xml}} some sub-topics in algebraic geometry and two textbooks. Our dataset is over five times larger and contains pairs extracted from advanced academic mathematical papers. 


There are multiple motivations for the design of the task and our dataset.
We believe it may help MIR by serving as a proxy for 
the search for the existence of a mathematical
result, or for theorems and proofs related to one another (e.g.\ using the same
proof technique), an important search tool for any digital mathematical library 
\cite{2014arXiv1404.1905M}.
Learning to match statements and proofs would also benefit computer-assisted 
theorem proving, as it is akin to tasks such as premise selection, 
also recently addressed with NLP methods \cite{DBLP:journals/corr/abs-1905-07961}.
\ignore{More generally, finding supporting information for or against a given statement,
is integral to tasks such as question answering
or fact-checking \cite{vlachos-riedel:2014:W14-25}.
Our mathematical statement-proof assignment task can be thought
of as the transposition of such problem to the very specific
domain of mathematical research articles.}

We provide first results on our proposed task with
an array of neural models, aimed at scoring the likelihood of relationship between a statement and its proof. An analysis through a \textbf{symbol replacement procedure} provides insight on what such neural models are capable of learning about mathematical equations and text.

We provide two methods for decoding, one is local decoding, matching a proof to a theorem in a greedy way, and one that provides a global bi-partite matching based on a structured max-margin objective.
Such an architecture may have applications to other
NLP problems that can be cast as maximum bipartite matching problems (for a recent similar use in a different context, see \citealt{shun2023erasure}).

Our analysis shows that pre-trained language models do not obtain significant ``mathematical insight'' for performing this matching, but rather rely on shallow matching. However, this does not prevent them from performing the matching relatively well in several carefully crafted scenarios, reaching an MRR of 73.7.

\section{Related Work}



Most NLP work on mathematical discourse focuses on improving
Mathematical Information Retrieval \cite[MIR]{DBLP:conf/ntcir/ZanibbiAKOTD16} by 
establishing connections between mathematical formulae and natural language text
in order to improve the representation of formulae.

The interpretation of variables is highly dependent on the context.
For example, the symbol $E$ could denote an expectation in a statistics article,
or the energy in a physics article.
Some studies use the surrounding context of a formula
to assign a definition or a type to the whole formula, or to specific variables.
\newcite{W10-3910} focus on identifying coreferences between mathematical
formulae and mathematical concepts in Wikipedia articles.
\newcite{Kristianto12extractingdefinitions} extract definitions of mathematical expressions.
\newcite{zbMATH05679807}, \newcite{wolska2011} and \newcite{Schubotz:2016:SIM:2911451.2911503}
disambiguate mathematical identifiers, such as variables, using the surrounding textual context.
\newcite{N18-1028} infer the type of a variable in a formula from the textual context of the formula.
Another line of work focused on identifying specialized terms or concepts
to improve MIR \cite{P15-2055,C16-1221}.

Some work adapts standard NLP tools to the specificity of mathematical discourse,
e.g.\ POS taggers \cite{DBLP:journals/corr/SchonebergS14},
with the objective of using linguistic features to improve the search for definitions 
of mathematical expressions \cite{DBLP:journals/corr/PagaelS14}.
More recent work focuses on typing variables in mathematical articles \cite{ferreira-etal-2022-integer}, modeling formulae \cite{tangent-cft,DADURE2021},
and selecting premises \cite{premise-ferreira-freitas-2020,star-ferreira-freitas-2021}. 

An earlier version of our work covers some of the material in this paper \cite{coavoux2021learning}. The main differences between that version and the current version are the introduction of the symbol replacement evaluation (\S\ref{section:symbol}) and the use of pre-trained language models rather than recurrent neural networks.

\section{Task Description}

Given a collection of mathematical statements $\{s^{(i)}\}_{i \leq N}$, and a separate equal-size collection of mathematical proofs $\{p^{(i)} \}_{i \leq N}$, we are interested in the problem of assigning a proof to each statement.




\paragraph{Evaluation} We use two evaluation metrics.
Assuming that a system predicts a ranking of proofs,
instead of providing only a single proof, we evaluate
its output with the Mean Reciprocal Rank (MRR) measure:
$\text{MRR}(\{\hat r_i \}_{i \in \{1, \dots N\}}) = \frac{1}{N} \sum_{i=1}^N \displaystyle\frac{1}{\hat r_i}$, 
where $N$ is the number of examples and $\hat r_i$
is the rank of the gold proof for statement number $i$,
as predicted by the system.

As a second evaluation metric, we use a simple accuracy,
i.e.\ the proportion of statements whose first-ranked proof
is correct.

By construction (see \S\ref{sec:data}), it is possible though unlikely that the same mathematical statement occurs several times in the dataset.
It is more unlikely that several occurrences have exactly the same formulation and use the same variable names. Therefore, we consider a match to be correct if and only if it is associated with its original proof.

\section{Dataset Construction}
\label{sec:data}
\label{section:data}

This section describes the construction of the \textsc{MATcH} dataset of statement-proof pairs (see Figure~\ref{fig:pair-example} for an example).

\paragraph{Source Corpus} 
We use the MREC
corpus\footnote{\url{https://mir.fi.muni.cz/MREC/}, version 2011.4.439.} \cite{dml:702604}
as a source.
The MREC corpus contains around 450k articles from ArxMLiV \cite{stamerjohanns2010transforming},
an on-going project aiming at converting the arXiv\footnote{\url{https://arxiv.org/}}
repository from \LaTeX~to XML, a format more suited to machine processing.
In this collection, mathematical formulae are represented in the
MathML\footnote{\url{https://www.w3.org/Math/}} format,
a markup language.



\paragraph{Statistics}
We extract statement-proof pairs as described in Appendix~\ref{sec:appendix:dataset}. Our processing of MREC includes the identification of statement-proof pairs through meta tags and the linearization of the representation of mathematical equations.

\begin{table}[t!]
    \resizebox{\columnwidth}{!}{
        \begin{tabular}{lrr}
            \toprule
            Number of articles in the MREC corpus          &&  439,423 \\
            Extracted articles with statement-proof pairs  &&   27,841 \\
            Total number of statement-proof pairs           &&  184,094 \\
            Number of (primary) categories            &&  (120) 135\\
            Average number of categories per article  &&    1.7\\
            \midrule
            \midrule
            Most represented primary categories & \# articles & \# \ pairs \\
            \midrule
math.AG Algebraic Geometry 		&	2848 &	22029 \\
math.DG Differential Geometry  &	2030 &	12440 \\
math.CO Combinatorics 			&	1705 &	10548 \\
math.GT Geometric Topology		&	1539 &	9234 \\
math.NT Number theory          &	1454 &	9521 \\
math.PR Probability				&	1422 &	7660 \\
math.AP Analysis of PDEs &	1386 &	6981 \\
math-ph Mathematical Physics    &	1249 &	6491 \\
math.FA Functional Analysis    &	1143 &	8011 \\
math.GR Group Theory &	970 &	7806 \\
math.DS Dynamical System &	961 &	6424 \\
math.QA	Quantum Algebra & 944	& 8074 \\
math.OA	Operator Algebras  & 923	&     8050 \\
            \bottomrule
        \end{tabular}
    }
    \caption{Statistics about the dataset and categories of mathematical articles.}
    \label{tab:statistics}
\end{table}

\begin{table}
    \resizebox{\columnwidth}{!}{
        \begin{tabular}{lrrr}
            \toprule
            Statements   & Min   & Max & Mean$\pm$SD \\
            \midrule
            Text+math  &    20  &  500    & 80$\pm$57 \\
            Text only  &    1   &  398    & 30$\pm$20 \\
            Math only  &    0   &  470    & 58$\pm$20 \\
            Math proportion & 0$\%$ & 99.5$\%$ & 58$\%\pm$20\\
            \midrule
            Proofs     & \\
            \midrule
            Text+math  & 20 & 500 & 210$\pm$ 127 \\
            Text only  & 1 & 467  & 81  $\pm$ 56 \\
            Math only  & 0 & 495  & 129 $\pm$ 96 \\
            Math proportion & 0$\%$ & 99.6$\%$  & 56$\%\pm$ 21 \\
            \bottomrule
        \end{tabular}
    }
    \caption{Number of tokens in the dataset. We report for statements
    and proofs the minimum, maximum and average number of tokens
    broken down by type (`math' for tokens extracted from
    formulae and `text' for the others).
    A value of 0 for, e.g.\ the `math only' row, means that the statement or proof does not contain mathematical symbols or formulae.}
    \label{tab:token-stats}
\end{table}

We report in Table~\ref{tab:statistics} some statistics about the dataset we collected.
The extracted articles were from a diverse set of mathematical subdomains,
and connected domains, such as computer science (746 articles from 30 subcategories)
and mathematical physics (2562 from 31 subcategories).
There are in average 6.6 statement-proof pairs per article. We report statistics about the size of statements
and proofs in the number of tokens in Table~\ref{tab:token-stats}.
We report the number of tokens in formulae (math), in the text itself (text) and in both (text+math).
On average, proofs are much longer than statements.
Statements and proofs have approximately the
same proportion of text and math.
Overall, the variation in the number of tokens across statements and
proofs is extremely high, as illustrated by the standard deviation (SD)
of all presented metrics.

\section{Symbol Replacements}
\label{section:symbol}

With our current dataset setup, we implicitly make the assumption that both the theorem and the proof are authored by the same authors. This assumption is incongruent with the MIR-flavor of our task. First, it is not useful for researchers to match proofs they authored. Second, each person has a unique writing style expressed by unique mathematical jargon and notations.
To relieve of this assumption, we introduce several symbol replacement levels for changing the names of the proof variables. Then, we train and test our models using these altered datasets. These replacement levels also provide insight on the ability of our models to semantically analyze the input statement-proof pairs.
\paragraph{Symbol Replacement Levels}
We propose different levels of symbol replacement, focusing on mathematical notation. More precisely, we aim to replace the proof variable names if they appear in the statement without damaging the proof semantics. To do that, we change symbols that appear both in the proof and the statement. We do not change constant symbols such as $\pi$, as they often carry semantic meaning outside of the proof scope. 

We experiment with four levels of symbol replacement (examples in parenthesis):

\begin{itemizesquish}{-0.3em}{0.5em}
    \item \textbf{Symbol conservation} - all symbols remain intact, so the theorem and the proof overlap. All previous work uses that. $\hfill$ ($a_n = a_{n-1} + a_{n-2}$)
    \item \textbf{Partial symbol replacement} - A fraction of $\alpha$  of all the symbols in the proof remain the same, and the rest are changed. In our experiments, we use $\alpha = 0.5$. $\hfill$ ($x_n = x_{n-1} + x_{n-2}$)
    \item \textbf{Full symbol replacement}  - all symbol names are changed ($\alpha = 1.0$ as above). $\hfill$ ($x_i = x_{i-1} + x_{i-2}$) 
    \item \textbf{Symbol transposition} - We permute the variables' names such that no symbol remains the same, thus changing their original functionality. $\hfill$ ($n_a = n_{a-1} + n_{a-2}$)
\end{itemizesquish}

More details about this appear in Appendix~\ref{sec:appendix:anonymization}.

%

\section{Bilinear Similarity Model}

We propose a model based on an encoder (\S\ref{sec:setup}) 
that constructs fixed-size vector representations for statements and proofs, and a similarity function that scores the relatedness
of a statement-proof pair.

\paragraph{Trainable Bilinear Similarity Function}
Given the encoded representations of a statement $\mathbf s = \text{enc}(s)$
and a proof $\mathbf p = \text{enc}(p)$, we compute an association score
with the following bilinear form:
\begin{equation*}
    \text{score}(\mathbf s, \mathbf p) = \mathbf s^{\top} \cdot \mathbf W \cdot \mathbf p + b,
\end{equation*}
where $\mathbf W$ and $b$ are parameters that are learned together with
a self-attentive encoder parameters (\S\ref{section:encoders}).

\paragraph{Local Decoding}
For a collection of $n$ statements and proofs, we first score
all possible pairs ($s, p$), and construct a matrix $M = (m_{ij}) \in \mathbb{R}^{n\times n}$,
with
\begin{equation*}
m_{ij} = \text{score}(\mathbf s^{(i)}, \mathbf p^{(j)}),
\end{equation*}
where $\mathbf s^{(i)}$ and  $\mathbf p^{(j)}$ are
the encoded representations of, respectively, 
the $i^{th}$ statement and the $j^{th}$ proof.
Then we can straightforwardly sort each row by decreasing order and assign
the proof ranking to the corresponding statement.
The best ranking proof $\hat p$ for statement $i$ satisfies
$\hat p^{(i)} = \argmax_{j} \; m_{ij}$.
We call this decoding method `local', since it does not take
into account dependencies between assignments.
In particular, several statements may have the same highest-ranking proof.

\paragraph{Global Decoding}
The local decoding method overlooks a crucial piece of information:
a proof should correspond to a single statement.
In a worst-case situation, a small number of proofs may score
high with most statements and be systematically assigned as
highest-ranking proof by the local decoding method.

In preliminary experiments, we analyzed the output
of our system with local decoding on the development set,
focusing on the distribution of the single highest-ranking proof for each statement.
We found that $23\%$ of the proofs were assigned to at least
two different statements, whereas more than $40\%$ of proofs
were assigned to no statement. See also Appendix~\ref{appendix:assign}.\footnote{We used a simple encoder for these experiments, which we describe in \S\ref{sec:setup} (NPT).}

We propose a second decoding method based on a global constraint on the output:
a proof can be assigned only to a single statement.
Intuitively, the constraint models the fact that 
if a proof is assigned by the system to a certain statement with high confidence,
we can rule it out as a candidate for other statements.
Under this constraint, the decoding problem reduces to
a classical maximum weighted bipartite matching problem,
or equivalently, a Linear Assignment Problem (LAP).
In more realistic scenarios (e.g.\ if the input sets of statements and proofs do not have the same size), the method would require some adaptation.

Formally, we define an assignment $A$ as a Boolean
matrix $A = (a_{ij})\in \{0,1\}^{n\times n}$ with the following constraints:
\begin{align*}
    \forall i \forall j, \sum_{j} a_{ij} = \sum_{i} a_{ij} = 1,
\end{align*}
i.e.\ each row and each column of $A$ contains a single non-zero coefficient.
The score of an assignment $A$ is the sum of scores of the chosen edges:
\begin{align*}
    \text{score}(A,M) = \sum_i \sum_j a_{ij} m_{ij}.
\end{align*}
Finally, global decoding consists in solving the following LAP:
\begin{align*}
    \hat A(M) = \argmax_{\substack{A \in \{ 0,1 \}^{n\times n} \\\text{s.t.\ } \forall i \forall j, \sum_{j} a_{ij} = \sum_{i} a_{ij} = 1}}  \text{score}(A, M).
\end{align*}

The LAP is solved in polynomial time by
the Hungarian algorithm \cite{Kuhn1955},
the LAP-Jonker-Volgenant algorithm \cite[LAP-JV;][]{Jonker1987},
or the push-relabel algorithm \cite{Goldberg1995}.
These methods have a $\mathcal{O}(n^3)$ time complexity where $n$
is the number of pairs, and $\mathcal{O}(n^2)$ memory complexity.
This is too expensive in our case, due to our dataset size.

To remedy this limitation, when we perform decoding on a large set,
we only consider the $k$ best-scoring
proofs (i.e.\ outgoing edges in the bipartite graph)
for each statement, which makes the number of edges linear
in the number of pairs $n$ (considering $k$ fixed).
Moreover, we use a modification of the LAP-JV algorithm specifically
designed for sparse matrices \cite[LAP-MOD;][]{VOLGENANT1996917}.






\section{Local and Global Training}

We propose two training methods for the similarity
model above:
a local training method that
only considers statements in isolation (\S\ref{sec:local})
and a global model trained to predict a bipartite matching
(\S\ref{sec:global}), with a hybrid global-local objective.

\subsection{Local Training}
\label{sec:local}

We would like to
train our model to assign a high similarity to the gold
statement-proof pair, and a low similarity to all other
statement-proof pairs.
This corresponds to the following objective, for a single
statement $s$ and its gold proof $p$:
\begin{align*}
    \mathcal{L}_{\textsc{loc}}(s, p, P; \boldsymbol \theta) &= - \log \mathbb P(p | s; \boldsymbol \theta) \\
                      &\!\!\!\!\!\!\!\!= - \log \left( 
\dfrac{\exp({\text{score}(\mathbf s, \mathbf p)})}{\sum\limits_{p' \in P} \exp({\text{score}(\mathbf s, \mathbf p')})}
\right),
\end{align*}
where $P$ is the set of proofs, and $\boldsymbol \theta$ are the parameters of the model.
Directly optimizing this loss function requires the computation
of $\mathbf p = \text{enc}(p)$ for every proof in the dataset,
for a single optimization step.
This is not realistic considering memory limitations,
the size of the train set,
and the fact that our self-attentive encoder is
the most computationally expensive part of the network.

Instead, we sample minibatches of $b$ pairs and optimize the following
proxy loss for the sequence $S'=(s_1, \dots, s_b)$ of statements
and the sequence $P'=(p_1, \dots, p_b)$ of corresponding proofs:\footnote{We also
experimented with a Noise-Contrastive Estimation approach \cite{journals/jmlr/GutmannH12}.
However, it exhibited a much slower convergence rate.}
\begin{equation*}
    \mathcal{L'}_{\textsc{loc}}(S', P'; \boldsymbol \theta) = \sum_{i=1}^b \mathcal{L}_{\textsc{loc}}(s^{(i)}, p^{(i)}, P'; \boldsymbol \theta).
\end{equation*}
In practice, we sample uniformly and without replacement $b$ pairs from 
the training set at each stochastic step.



\subsection{Hybrid Local and Global Training}
\label{sec:global}

The local training method only considers statements in isolation.
Even though we expect a locally trained model to perform better
with global decoding, we hypothesize that a model that is trained
to predict the full structure (a bipartite matching)
will be even better.

For a collection of $n$ proofs and $n$ statements,
the size of the search space (i.e.\ the number of bipartite
matchings) is $n!$, since each matching corresponds
to a permutation of proofs.
As a result, the use of a globally normalized model
is impractical.
We turn to a max-margin model that does not
require normalization over the full search space.

We use the following max-margin objective, for a set $B$ of $n$ pairs corresponding to matrix $M$:
\begin{align*}
    \mathcal{L}_{\textsc{glob}}(B;\boldsymbol \theta) = & \max(0, \Delta(\hat A, I) \\ & +  \text{score}(\hat A, M) - \text{score}(I, M)),
\end{align*}
where $\boldsymbol \theta$ is the set of all parameters
$\hat A$ is the predicted assignment and $I$ is
the gold assignment, i.e.\ the identity matrix.
The structured cost 
\begin{equation*}
\Delta(\hat A, I) = \sum_{ij} \max(0, (\hat A - I)_{ij})
\end{equation*}
aims at enforcing a margin for each individual assignment.

The computation of this loss requires exact decoding for each optimization step.
Since exact decoding is only feasible for a small $n$,
and since we need to keep track of all intermediary
vectors to compute the backpropagation step,\footnote{In particular,
the computation graph needs to conserve all encoding
layers for the $2n$ texts involved.}
we perform each stochastic optimization step on a minibatch
of pairs of size $b$.
Since this global objective had a slow convergence rate (\S\ref{sec:setup}),
in practice, we use a hybrid local-global objective: $\mathcal{L'}_{\textsc{loc}} + \mathcal{L}_{\textsc{glob}}$.

\section{Experimental setup}
\label{sec:setup}
\label{sec:results}
\begin{table*}[htp]
\centering
\begin{tabular}{l|cccccccc}
\hline
                   & \multicolumn{8}{c}{Symbol Replacement Level}                                                                                \\
                   & \multicolumn{2}{c|}{Conservation} & \multicolumn{2}{c|}{Partial} & \multicolumn{2}{c|}{Full} & \multicolumn{2}{c}{Transposition} \\ \hline
Encoder-Decoder    & MRR         & Acc         & MRR           & Acc          & MRR         & Acc         & MRR            & Acc            \\ \hline
NPT-Local-Local     & 63.22       & 56.08       & 47.19         & 39.24        & 40.36       & 32.52       & 56.17          & 48.30          \\
NPT-Local-Global    & -           & 61.89       & -             & 42.55        & -           & 35.43       & -              & 53.49          \\
NPT-Global-Global   & -           & 62.14       & -             & 43.68        & -           & 35.85       & -              & 55.28          \\
\textsc{ScratchBERT}-Local-Local   & \textbf{73.73}       & 67.12       & \textbf{64.79}         & 57.20        & \textbf{60.67}       & 52.54       & \textbf{73.17}          & 66.51          \\
\textsc{ScratchBERT}-Local-Global  & -           & \textbf{74.68}       & -             & \textbf{62.80}        & -           & \textbf{57.69}       & -              & \textbf{74.03}          \\
\textsc{ScratchBERT}-Global-Global & -           & 71.38       & -             & 58.06        & -           & 52.31       & -              & 70.32          \\
\textsc{MathBERT}-Local-Local   & 54.51       & 46.45       & 44.31         & 36.10        & 38.91       & 30.62       & 52.57          & 44.52          \\
\textsc{MathBERT}-Local-Global   & -           & 49.77       & -             & 37.92        & -           & 32.03       & -              & 47.43          \\
\textsc{MathBERT}-Global-Global & -           & 45.38       & -             & 33.64        & -           & 28.47       & -              & 43.41         
\end{tabular}
\caption{The MRR and accuracy scores for different combinations of encoders, decoders, and symbol replacement levels. All the models are trained and tested on the same replacement level. Best result in each column is in bold. Following the model name, we include its encoder and decoder type (both being either Local or Global). We do not include MRR scores for global inference, as there matching is done for all theorems together without ranking.}
\label{tab:main-results}
\end{table*}

\paragraph{Dataset} We use the dataset whose construction
is described in \S\ref{sec:data}.
We shuffle the collection of statement-proof pairs before
performing a $80\%/10\%/10\%$ train-development-test split,
corresponding to 147278 pairs for the training sets and 18408 pairs
for the development and tests. We experiment with a default, Mixed, split and a harder Unmixed split (see \S\ref{section:cross}).


\paragraph{Encoders}
\label{section:encoders}

We experiment with several encoders to obtain neural representations of the theorem and proof pairs. 
Our first encoder is a simple self-attentive encoder. We use $\ell=2$ self-attentive layers with 4~heads to obtain contextualized embeddings of dimension $d=300$. The query and key vectors have size $d_k=128$. We construct a vector representation for the text with a max-pooling
layer over the contextualized embeddings of the last self-attention layer. We do not use any form of pre-training for this encoder and hence name it ``no pre-training encoder'' (\textsc{NPT}). In addition, we experiment with a BERT model \cite{devlin-etal-2019-bert} as an encoder. We do not use the pre-trained version provided by \citeauthor{devlin-etal-2019-bert}, but rather pre-train the base version from scratch (\textsc{ScratchBERT}), but we do compare our results against a math-tailored pre-trained version of BERT (\citealt{peng2021mathbert}; see below). Both the \textsc{NPT} and \textsc{ScratchBERT} vocabularies are customized for our dataset, as preliminary experiments  revealed the importance of the model-task vocabulary match.\footnote{This supports the findings of \newcite{chalkidis-etal-2020-legal}, for example, in a different domain.} 

To further demonstrate how crucial this vocabulary match is, we experiment with math-BERT (\textsc{MathBERT}; \citealt{mathbert}), a state-of-the-art pre-trained model for mathematical formula understanding. This model is pre-trained on a large mathematical corpus ranging from pre-kindergarten, to high-school, to college graduate level mathematical content, including professional mathematical papers, using the BERT masked language modeling (MLM) task. We use the pre-trained version provided by the authors, \emph{without vocabulary customization}. All of our encoders are fine-tuned on the matching task. In addition, we experiment with a naive token-matching system that computes cosine similarities between TF-IDF representations of statements and proofs. We discovered that their performance was very low, ranging from 11.4 to 29.8 (MRR), so we did not experiment with them further.



%

\paragraph{Hyperparameters}

For pretraining \textsc{ScratchBERT}, we first train a new word piece tokenizer\footnote{\url{https://huggingface.co/docs/transformers/tokenizer_summary\#wordpiece}}. Next, we train the \textsc{ScratchBERT} model on the MLM task for 60 epochs (around 3 days) using four NVIDIA V100 GPUs. We evaluate the language model every 500 steps, where one step stands for training on one example, and choose the one with the best performance on the validation set.

We perform local and global training / finetuning respectively for the NPT model, \textsc{MathBERT}, and \textsc{ScratchBERT}. NPT has 15M parameters while \textsc{MathBERT} and \textsc{ScratchBERT} have 110M parameters. We observed in initial experiments that training only with the global objective required a long time to converge. Therefore, we used the following global-local objective: $\mathcal{L'}_{\textsc{loc}} + \mathcal{L}_{\textsc{glob}}$, that we optimized by alternating one stochastic step for each loss.

We train the NPT model for 400 epochs (around 1 day with two GPUs) over the whole training set for local and global training. We use batches of size $b=60$ and set learning rate $l=5\times 10^{-3}$ with the Averaged Stochastic Gradient Descent (ASGD; \citealt{asgd}) optimizer. We use an exponential learning rate scheduler (the learning rate multiplied by $0.996$ after each epoch) to stabilize the optimizer in the latter training procedure (after 300 epochs). We evaluate the performance of the model  on the validation set every 20 epochs during training and select the best one among these intermediate models. 

We use four NVIDIA V100 GPUs to fine-tune \textsc{MathBERT} and \textsc{ScratchBERT} on the training set for 60 epochs (around 2 days) with a learning rate of $l=2\times 10^{-3}$, an ASGD optimizer, batches of size $b=16$, and a scheduler that multiplies the learning rate by $0.99$ after each epoch. 
We choose the best model on the validation set, evaluating the models
every five epochs.

\paragraph{Global Decoding}
Recall that exact global decoding is only feasible for a small subset of pairs.
During global training, we chose a batch size small enough to perform exact decoding.
However, it is not feasible to perform exact decoding on the whole development and test corpora.
Therefore, we prune the search space by keeping only the 500-best
candidate proofs for each statement, and use the LAP-MOD algorithm designed for sparse matrices.
In practice, we used the implementations of the LAP-JV and LAP-MOD algorithms
from the \texttt{lap} Python package,\footnote{\url{https://github.com/gatagat/lap}}
for respectively exact decoding on mini-batches during global training
and decoding on whole datasets during evaluation.

\section{Results}
First, we assess the task difficulty under different replacement levels using different encoders and schemes (global or local training, global or local decoding). In particular, we are interested in assessing
whether global decoding improves accuracy when training
is only local, and how the more complex global training
method fares with respect to local training.
We then measure the informativeness
of different types of input: text, mathematical formulae, or both. The comparison of these settings is meant to provide insight
into which type of information is crucial to the task.
Finally, we experiment with a cross replacement levels setup, i.e., when a model is tested on a different symbol replacement level from the one that was used during training. We hope this experiment will shed some light on the importance of training models on real-world datasets. 
\subsection{Main Results}
\begin{table*}[htp]
\centering
\begin{tabular}{cl|cccccccc}
\hline
 & \multirow{3}{*}{\backslashbox{Source}{Target}} & \multicolumn{8}{c}{Symbol Replacement}                                                                                \\
                        &    & \multicolumn{2}{c|}{Conservation} & \multicolumn{2}{c|}{Partial} & \multicolumn{2}{c|}{Full} & \multicolumn{2}{c}{Transposition} \\\hhline{~~--------}
        &                      & MRR         & Acc         & MRR           & Acc          & MRR         & Acc         & MRR            & Acc            \\
                              \hline
 \multirow{4}{*}{\rotatebox{90}{Mixed}}  & Conservation                         & 73.73       & 67.12       & 43.87         & 36.36        & 29.74       & 25.36       & 69.56          & 62.23          \\
& Partial                      & \textbf{74.21}       & \textbf{67.96}       & \textbf{64.79}         & \textbf{57.20}        & 53.77       & 45.40       & 72.13          & 65.42          \\
& Full                         & 65.26       & 57.63       & 63.01         & 55.13        & \textbf{60.67}       & \textbf{52.54}       & 64.59          & 56.92          \\
& Transposition                  & 73.78       & 67.40       & 43.67         & 36.02        & 29.76       & 25.47       & \textbf{73.17}          & \textbf{66.51}\\   
\hline
\hline
\multirow{4}{*}{\rotatebox{90}{Unmixed}}  & Conservation                         & \textbf{67.62}       & 57.54       & 21.26         & 13.83        & 7.09       & 3.68       & 59.54          & 48.61          \\
& Partial                      & 61.19       & 50.73       & \textbf{55.26}         & \textbf{44.45}        & 50.68       & 39.94       & 59.63          & 49.01          \\
& Full                         & 55.68       & 45.18       & 54.92         & 44.34        & \textbf{54.62}       & \textbf{44.22}       & 55.38          & 44.91          \\
& Transposition                  & 67.5      & \textbf{57.76}       & 23.31         & 15.26        & 8.98       & 4.97       & \textbf{66.25}          & \textbf{59.29}\\   
\hline

\end{tabular}%
\caption{Cross-replacement levels performance for the \textsc{ScratchBERT}-Local-Local model for both splits: Mixed and Unmixed.}
\label{tab:anonymized-results}
\end{table*}
Table~\ref{tab:main-results} presents our results. We report MRR (if relevant) and accuracy scores across different levels of symbol replacement. 

\begin{table}[htp]
\centering
\begin{tabular}{c|cccc}
\hline
      & \multicolumn{4}{c}{Symbol Replacement}              \\
      & \multicolumn{2}{c|}{Conservation} & \multicolumn{2}{c}{Full} \\ \hhline{~----}
Input & MRR         & Acc         & MRR         & Acc        \\ \hhline{~----}
      & \multicolumn{4}{c}{NPT}                              \\ \hline
Text  & 22.51       & 16.68       & 22.51       & 16.68      \\
Math  & 65.08       & 58.47       & 34.55       & 27.30      \\
Both  & 63.22       & 56.08       & 40.36       & 32.52      \\
\hline
      & \multicolumn{4}{c}{\textsc{ScratchBERT}}                            \\ \hline
Text  & 36.85       & 29.18       & 36.85       & 29.18      \\
Math  & 63.10       & 55.92       & 41.64       & 34.01     \\
Both  & \textbf{73.73}     & \textbf{67.12}      & \textbf{60.67}       & \textbf{52.54}     
\end{tabular}
\caption{\textsc{ScratchBERT}-Local-Local and NPT-Local-Local performance for different input types. Both stand for the original and complete input.}
\label{tab:input-results}
\end{table}

\paragraph{Encoders}
While \textsc{MathBERT} is pre-trained on millions of examples curated from mathematical contents, it performs worse than the less complex NPT encoder, which is trained solely on the downstream task across all symbol replacement levels and decoders.\footnote{We observe similar trends when fine-tuning the out-of-the-box BERT model on the matching task.} \textsc{ScratchBERT}, which shares \textsc{MathBERT} architecture and NPT customized vocabulary, is outperforming both consistently. These results demonstrate the vocabulary importance for learning from mathematical texts.  

\paragraph{Symbol Replacement Levels Difficulty}
Best performance is achieved when no symbol is replaced (Conservation), as the models can match identical symbols across theorem-proof pairs.
The models achieve similar performance with Transposition replacement. These results suggest that the symbols' order, context, and function within the mathematical text do not play a significant role when the theorem and proof share the same symbols. In contrast, when the symbol names are changed (Partial and Full replacements), we observe a sharp decline in results. 

\paragraph{Training and Decoding Effects}
In all settings, global \emph{decoding} substantially improves accuracy. These improvements are more noticeable for the NPT and \textsc{ScratchBERT} encoders. For NPT, we observe better performance when using global \emph{training}, but not for \textsc{ScratchBERT} and \textsc{MathBERT}. Due to the lack of computational resources, we can not reach the global training full potential when using highly expressive encoders such as \textsc{ScratchBERT} and \textsc{MathBERT}, which share BERT-base architecture. 

\subsection{Effect of Input Type Analysis}
To better understand the importance of each input type, we examine \textsc{ScratchBERT}-Local-Local and NPT-Local-Local performance when fed with text, mathematical formulae, or both (Table~\ref{tab:input-results}). We test them on the Conservation and Full symbol replacement levels. 
The mathematical formulae input plays a more significant role for both models than the textual input. When trained and tested on the Conservation replacement level, NPT-Local-Local makes better use of the mathematical formulae input than the more expressive, pre-trained \textsc{ScratchBERT}-Local-Local. When trained and tested with Full replacement, where the models cannot rely on simple token-matching, NPT-Local-Local suffers from a sharper performance decline than \textsc{ScratchBERT}-Local-Local when fed with mathematical formulae input. These results suggest that when applied to the Conservation data, a less expressive model can get high results by harnessing simple token matching. \textsc{ScratchBERT}-Local-Local performs better for both replacement levels when fed with text and complete input.

\subsection{Cross Replacement Setup}
\label{section:cross}

Table \ref{tab:anonymized-results} shows the effect of testing a model on different symbol replacements than the one the model was trained on.
We use the \textsc{ScratchBERT}-Local-Local model for all of our experiments. 
We observe a sharp decline in results when \textsc{ScratchBERT}-Local-Local is trained with Conservation and tested on Partial or Full. These drops in performance suggest the model developed a strong dependency on exact symbol name matching. In addition, the replacement shift from Conservation to Transposition and vice versa resulted in a minor performance drop. These results provide additional evidence for the lack of importance of mathematical functionality, order, and context of symbols' names shared across theorem and proof pairs. The model trained on the Partial symbol replacement level demonstrated significant resilience when tested with other symbol replacement levels. It outperforms the rest of the models when applied to out-of-domain replacement levels and the Conservation replacement level in-domain model.

In addition, we experimented with theorem-proof pairs split where pairs from the same paper could not appear in the same set: train, validation or test (Unmixed). All models exhibited a reduction in performance when trained and tested under these conditions. Particularly noteworthy was the decrease in performance observed in models that were trained using the Conservation and Transposition symbol replacement methods and evaluated on data using the Partial or Full replacement methods. These sharp declines highlight the dependence of the model on simple symbol matching rather than deeper inferential analysis.

\subsection{Protected Symbols}
\label{section:protected}

Insofar, we overlooked that some symbols carry a default meaning over a whole mathematical domain (\textbf{protected} symbols, e.g. $P(x)$ for probability). Replacing them locally may result in a detrimental impact on semantic mathematical content. We test the impact of substituting protected symbols in a controlled setting by comparing models trained with symbol replacement methods that preserve protected symbols versus methods that treat all symbols equally. The test set preserves the protected symbols. We follow the Unmixed setup, where theorem-proof pairs from the same paper must appear in the same split. 

We focus on the probability theory domain. Focusing on a single domain enables us to construct a list of protected symbols more precisely. Our list consists of the $P$ (probability measure), $E$ (expected value), $V$ (variance), $\sigma$ (standard deviation and covariance), and $\rho$ (correlation) symbols.\footnote{We relied on Wikipedia, \url{https://tinyurl.com/2c3kwsfx}, for creating the protected symbols list.} Table~\ref{tab:CCR} shows that training the \textsc{ScratchBERT}-Local-Local model using the Partial+P replacement method results in slightly better results. We present only a subset of our results for brevity; the pattern re-occurs with all symbol replacement methods.

\begin{table}[htp]
\centering
\footnotesize
\begin{tabular}{l|cccccccc}

\hline
\multirow{3}{*}{\backslashbox{Source}{Target}} & \multicolumn{4}{c}{Symbol Replacement}                                                                                \\
                             & \multicolumn{2}{c|}{Conservation} & \multicolumn{2}{c}{Partial+P}  \\\hhline{~--------}
                             & MRR         & Acc         & MRR           & Acc                     \\
                              \hline
Conservation                         & \textbf{69.26}       & \textbf{59.59}       & 27.9         & 18.29                  \\
Partial                      & 61.36       & 51.72       & 54.06        & 42.67             \\
Partial+P                         & 62.1       & 51.92       & \textbf{55.92}         & \textbf{45.23}    
\\
Full                        & 53.63       & 42.08       &52.85         & 41.4 
\\
Full+P                         & 56.27       & 45.13       & \textbf{55.92}     &    44.84
\end{tabular}%
\caption{Controlled cross-replacement levels performance for the \textsc{ScratchBERT}-Local-Local model. Both train and test sets are curated from the probability theory domain. +P next to a symbol replacement method means that Protected symbols are not being replaced.}
\label{tab:CCR}
\end{table}

\subsection{Qualitative Analysis}

To study which tokens affect our model predictions, we use LIME \cite{ribeiro2016should}, a method for calculating feature importance. We examine \textsc{ScratchBERT}-Local-Local trained with the Conservation setup and with Full replacement. Both are applied to original test examples. We observe that the Conservation \textsc{ScratchBERT}-Local-Local model heavily relies on the mathematical tokens and barely benefits from the text ones. In contrast, the \textsc{ScratchBERT}-Local-Local model that was trained in the full symbol replacement setup strongly relies on textual tokens with mathematical meaning, such as \emph{module}, \emph{supplement}, and \emph{semistable}. We visualize that in Figure \ref{fig:qual}.

\section{Conclusion}

We developed a bilinear similarity model and a large dataset (\textsc{MATcH}) for a task focusing on the domain of mathematical research articles.
The task consists in matching a proof to a mathematical statement. We proposed two ways to train and inference with our model and dataset: local matching and global matching. We assessed the difficulty of the task with several pre-trained encoders,
demonstrating the importance of the vocabulary support for these models. Further assessment relies on using a symbol replacement procedure,
which helps test the type of mathematical reasoning the encoders can perform. While our model performs well on this task, we observe through the symbol replacement procedure
that the model makes a relatively shallow use of the text and formulae to obtain this performance.
\ignore{
Finally, we have introduced a global neural model for addressing
the structured prediction problem of maximum
weighted bipartite matching.
The model is based on a self-attentive encoder and a bilinear
similarity function.
Our experiments show that bag-of-words baselines are insufficient to solve the task,
and are outperformed by our proposed model by a wide margin.
We found that decoding is crucial to achieve
high results, and is further enhanced by a global training loss.
Finally, our results show that mathematical formulae are the
most informative source of information for the task but are best taken into account with the self-attentive neural model.
}

\section*{Limitations}
\label{sec:limitations}
Our work has three main limitations. 
First, we aim to simulate a setup where the same author did not write both a theorem and its corresponding proof. We reduce the intersection size of symbols between the statement and the proof, which leads to more challenging setups. In practice, authors and mathematical communities within fields differ in their use of notation and their writing style (creating \emph{mathematical language dialects}). Such overall dialect cannot be altered using simple rule-based methods. We leave it for future work to explore a full MIR setup for our task that takes this into consideration.

Second, due to computational limitations, we could not explore the full potential of our global training method. Our GPUs cannot handle large batch sizes for large models such as \textsc{MathBERT} and \textsc{ScratchBERT}. We use NVIDIA V100 GPUs that allow us to experiment with a batch size of 16 for \textsc{MathBERT} and \textsc{ScratchBERT}, compared to 60 with NPT. 

Third, while our symbol replacement method provides a coarse way to test the language model use of the symbols and text in mathematical articles, it presents cases in which the replacement is not precise. These cases arise because the use of symbols in mathematical language is rich and context-dependent (for example, while $\pi$ often refers to the pie constant, it might also refer to a tuple-projection function or a permutation). We partially address that in \S\ref{section:protected}.

\section*{Acknowledgments}

We thank the reviewers for their helpful comments, and Marcio Fonseca for feedback on an earlier draft. We also thank Richard Zanibbi for a discussion about aspects of this work. The experiments in this paper were supported by a compute grant from the Edinburgh Parallel Computing Center (Cirrus) and another compute grant for the Baskerville service at the University of Birmingham.

\bibliography{anthology,arr}


\appendix

\section{Details on Dataset Construction}
\label{sec:appendix:dataset}

As mentioned in \S\ref{section:data}, we use the MREC corpus to extract statement-proof pairs.

\paragraph{Statement-proof Identification}

For each XML article (corresponding to a single arXiv article),
we extract pairs of consecutive \texttt{<div>} tags such that:
    (i) the \texttt{class} attribute of the first \texttt{div} node
    contains the string \texttt{"theorem"};
    (ii) the \texttt{class} attribute of the second \texttt{div} node
    is the string \texttt{"proof"}.
Articles that do not contain such pairs of tags are discarded,
as well as articles that are not written in English
(representing 143 articles in French,
11 in Russian, 5 in German, 2 in Portuguese and 1 in Ukrainian),
as identified by the \texttt{polyglot} Python package.\footnote{\url{www.github.com/aboSamoor/polyglot/}}

In the remaining collection of pairs of statements and proofs,
we filter out pairs for which either the statement or the proof is too short.\footnote{We used a minimum length
of 20 tokens for both statements and proofs, based on a manual inspection of the shortest examples. We also exclude proofs and statements longer than 500 tokens.}
Indeed, the short texts were often empty (only consisting of a title, e.g.\ ``5.26 Lemma.''),
which we attribute to the noise inherent to the conversion to XML,
or not self-contained.
In particular, we identified several prototypical cases:
\begin{itemize}[noitemsep]
 \item Omitted (or easy) proofs contain usually a single word
 (`omitted', `straightforward',
 `well-known',
 `trivial',
 `evident'),
 but are sometimes more verbose (`This is obvious and will be left to the readers').
 \item Proofs that consist of a single reference to
 \begin{itemize}[noitemsep]
 \item An appendix (`See Appendix A');
 \item Another theorem (`This follows immediately from Proposition 4.4 (ii).');
 \item The proof method of another theorem (`Similar to proof of Lemma 6.1')
 \item Another article (`See [BK3, Theorem 4.8].');
 \item Another part of the article
 (`The proof will appear elsewhere.', `See above.', `Will be given in section 5.').
 \end{itemize}
\end{itemize}
Filtering on the number of tokens also excludes self-contained short proofs,
such as `Take $Q^\prime = p h_i - p_i$.'
However, such proofs were very infrequent on manual inspection
of the discarded pairs (2 in a manually inspected random sample of~100 discarded proofs).

\paragraph{Preprocessing: Linearizing Equations}
Mathematical formulae in the XML articles are enclosed
in a \texttt{<math>} markup tag, that materializes the switch to the MathML format,
and whose internal structure represents the formula as an XML tree.
As a preprocessing step, we linearize each formula to a raw sequence of strings.

In MathML, an equation can be encoded in a content-based (semantic) way or in a presentational way,
using different sets of markup tags.
We first convert all MathML trees to presentational MathML
using the XSL stylesheet from the Content MathML Polyfill repository.\footnote{\url{https://github.com/fred-wang/webextension-content-mathml-polyfill}}
Then we perform a depth-first search on each tree rooted in a \texttt{<math>} tag
to extract the text content of the whole tree.

During this preprocessing, we tested several processing choices:
\begin{itemize}[noitemsep]
 \item \textbf{Font information}. In mathematical discourses, fonts play an important role.
  Their semantics depend on conventions shared by researchers.
  If both $x$ and $\mathbf x$ appear in the same article,
  they are most likely to represent different mathematical objects, e.g.\ a scalar and a vector.
  Therefore, we use distinct symbols for tokens that are in distinct fonts.
  \item \textbf{Math-English ambiguity}. Some symbols can be used
  both in natural language text and in formulae. 
  For example, `a' can be a determiner in English, or a variable name in a formula.
  To avoid increasing ambiguity when linearizing formula, we type each symbol (as math or text)
  to make the mathematical vocabulary completely disjoint from the text vocabulary.
\end{itemize}
Both these preprocessing steps had a beneficial effect on the baselines in preliminary experiments.

\section{Distribution of Proof-Statement Assignments}
\label{appendix:assign}
\label{appendix:assignment}

Table~\ref{tab:cumulative} depicts the cumulative distribution of proofs and the number of statements they are assigned to.

\begin{table}
    \begin{center}
    \begin{tabular}{lrr}
        \toprule
        Statements & Proofs & $\%$\\
        \midrule
        $\geq 20$  & 7 & 0.0 \\
        $\geq 10$  & 80 & 0.2 \\
        $\geq 5$  & 1027 & 1.9 \\
        $\geq 2$  & 11949 & 22.6 \\
        $=1$  & 19531 & 37.0 \\
        $<1$  & 21275 & 40.3 \\
        \bottomrule
    \end{tabular}
    \end{center}
    \caption{Cumulative distribution of proofs in the development set,
    by number of statements to which they are assigned
    with the local decoding method.}
    \label{tab:cumulative}
\end{table}

\begin{figure*}[ht]
\begin{subfigure}[b]{0.48\textwidth}
  \makenonemptybox{5cm}{
  \parbox{\textwidth}{
    \textbf{Lemma 3.2.} Let \colorbox[HTML]{ffb26e}{$M$} be \colorbox[HTML]{ffbc81}{a} module \colorbox[HTML]{ffc28c}{and} \colorbox[HTML]{ffb26e}{$H$} \colorbox[HTML]{ffbc81}{a} local submodule of \colorbox[HTML]{ffb26e}{$M$}. Then \colorbox[HTML]{ffb26e}{$H$} \colorbox[HTML]{ffb472}{is} \colorbox[HTML]{ffbc81}{a} \colorbox[HTML]{ffb675}{supplement} of each proper submodule \colorbox[HTML]{ffb472}{$K$} $\leq$ \colorbox[HTML]{ffb26e}{$M$} with \colorbox[HTML]{ffb26e}{$H$} $+$ \colorbox[HTML]{ffb472}{$K$} $=$ \colorbox[HTML]{ffb26e}{$M$}.
    \\
    \textbf{Proof.} Since \colorbox[HTML]{ffb472}{$K$} \colorbox[HTML]{ffb472}{is} \colorbox[HTML]{ffbc81}{a} proper submodule of \colorbox[HTML]{ffb26e}{$M$} and \colorbox[HTML]{ffb472}{$K$} $+$ \colorbox[HTML]{ffb26e}{$H$} $=$ \colorbox[HTML]{ffb26e}{$M$}, \colorbox[HTML]{1f77b4}{we} \colorbox[HTML]{ffbe85}{have} \colorbox[HTML]{ffb472}{$K$} $\cap$ \colorbox[HTML]{ffb26e}{$H$} \colorbox[HTML]{ffb472}{is} \colorbox[HTML]{ffbc81}{a} proper submodule of \colorbox[HTML]{ffb26e}{$H$}. Therefore \colorbox[HTML]{ffb472}{$K$} $\cap$ \colorbox[HTML]{ffb26e}{$H$} $\ll$ \colorbox[HTML]{ffb26e}{$H$}, since \colorbox[HTML]{ffb26e}{$H$} \colorbox[HTML]{ffb472}{is} local. That \colorbox[HTML]{ffb472}{is}, \colorbox[HTML]{ffb26e}{$H$} \colorbox[HTML]{ffb472}{is} \colorbox[HTML]{ffbc81}{a} \colorbox[HTML]{ffb675}{supplement} of \colorbox[HTML]{ffb472}{$K$} in \colorbox[HTML]{ffb26e}{$M$}.
    \\(\url{https://arxiv.org/pdf/0810.0041.pdf})
    } 
  }
  \caption{Example statement/proof 1 - Symbol conservation}
  \label{fig:zero-a}
\end{subfigure}
\hfill
\begin{subfigure}[b]{0.48\textwidth}
  \makenonemptybox{5cm}{
  \parbox{\textwidth}{
    \textbf{Lemma 3.2.} Let $M$ be a \colorbox[HTML]{ffbd83}{module} \colorbox[HTML]{ff9333}{and} $H$ a \colorbox[HTML]{ffa951}{local} \colorbox[HTML]{ffa75a}{submodule} of $M$. Then $H$ \colorbox[HTML]{ff9a41}{is} a \colorbox[HTML]{ff8a22}{supplement} of each proper \colorbox[HTML]{ffa75a}{submodule} $K$ $\leq$ $M$ with $H+K=M$.
    \\
    \textbf{Proof.} Since $K$ \colorbox[HTML]{ff9a41}{is} a proper \colorbox[HTML]{ffa75a}{submodule} of $M$ \colorbox[HTML]{ff9333}{and} $K+H=M$, \colorbox[HTML]{2379b5}{we} \colorbox[HTML]{ffa658}{have} $K \cap H$ \colorbox[HTML]{ff9a41}{is} a proper \colorbox[HTML]{ffa75a}{submodule} of $H$. \colorbox[HTML]{1f77b4}{Therefore} $K \cap H \ll H$, since $H$ \colorbox[HTML]{ff9a41}{is} \colorbox[HTML]{ffa951}{local}. \colorbox[HTML]{93bedb}{That} \colorbox[HTML]{ff9a41}{is}, $H$ \colorbox[HTML]{ff9a41}{is} a \colorbox[HTML]{ff8a22}{supplement} of $K$ in $M$.
    \\ (\url{https://arxiv.org/pdf/0810.0041.pdf})
    } 
  }
  \caption{Example statement/proof 1 - Full symbol replacement}
  \label{fig:full-a}
\end{subfigure}
\\

\begin{subfigure}[b]{0.48\textwidth}
  \makenonemptybox{5cm}{
  \parbox{\textwidth}{
    \textbf{Lemma 4.1.9.} If \colorbox[HTML]{ff7f0e}{$\mathscr{F}$} \colorbox[HTML]{ffb778}{is} \colorbox[HTML]{bcd6e9}{a} \colorbox[HTML]{ff9b43}{$\mu$}-\colorbox[HTML]{ffc18a}{semistable} \colorbox[HTML]{c9deed}{$\mathscr{X}$}-twisted sheaf of \colorbox[HTML]{c5dcec}{rank} $r$ \colorbox[HTML]{d2e4f0}{then} dim Hom $($ \colorbox[HTML]{ff7f0e}{$\mathscr{F}$} $,$ \colorbox[HTML]{ff7f0e}{$\mathscr{F}$} $)$ $\leq$ $r^2$.
    \\
    \textbf{Proof.} Any endomorphism of \colorbox[HTML]{ff7f0e}{$\mathscr{F}$} \colorbox[HTML]{d1e3f0}{must} preserve the socle (see Lemma 1.5.5ff of [4]); moreover, the quotient \colorbox[HTML]{ff7f0e}{$\mathscr{F}$} $/$ Soc(\colorbox[HTML]{ff7f0e}{$\mathscr{F}$}) \colorbox[HTML]{ffb778}{is} also \colorbox[HTML]{ffc18a}{semistable}. The result follows by induction from the polystable case, which itself follows immediately from the fact that stable sheaves are simple.
    \\ (\url{https://arxiv.org/pdf/0803.3332.pdf})
    } 
  }
  \caption{Example statement/proof 2 - Symbol conservation}
  \label{fig:zero-b}
\end{subfigure}
\hfill
\begin{subfigure}[b]{0.48\textwidth}
  \makenonemptybox{5cm}{
  \parbox{\textwidth}{
    \textbf{Lemma \colorbox[HTML]{a8cae2}{4}.1.9.} \colorbox[HTML]{ffb06a}{If} $\mathscr{F}$ \colorbox[HTML]{ff9333}{is} a $\mu$-\colorbox[HTML]{ff7f0e}{semistable} $\mathscr{X}$-twisted sheaf \colorbox[HTML]{9dc4de}{of} \colorbox[HTML]{ffc897}{rank} \colorbox[HTML]{98c0dc}{$r$} then \colorbox[HTML]{ff8315}{dim} Hom $($ $\mathscr{F}$ $,$ $\mathscr{F}$ $)$ $\leq$ \colorbox[HTML]{98c0dc}{$r$}$^2$.
    \\
    \textbf{Proof.} Any endomorphism \colorbox[HTML]{9dc4de}{of} $\mathscr{F}$ \colorbox[HTML]{a1c6df}{must} preserve the socle (see Lemma 1.5.5ff \colorbox[HTML]{9dc4de}{of} [\colorbox[HTML]{a8cae2}{4}]); moreover, the quotient $\mathscr{F}$ $/$ Soc($\mathscr{F}$) \colorbox[HTML]{ff9333}{is} also \colorbox[HTML]{ff7f0e}{semistable}. The result follows by induction from the polystable case, which itself follows immediately from the fact that stable sheaves are simple.
    \\ (\url{https://arxiv.org/pdf/0803.3332.pdf})
    } 
  }
  \caption{Example statement/proof 2 - Full symbol replacement}
  \label{fig:full-b}
\end{subfigure}
\caption{LIME visualizations for the model that was trained in the symbol conservation setup (a and c) and their corresponding LIME visualizations for the model that was trained in the full symbol replacement setup (b and d). The LIME ``match'' class supporting features are colored in orange, and the ``mismatch'' is in blue. The darker the color, the higher (in absolute value) the feature importance.}
\label{fig:qual}
\end{figure*}

\section{Symbol Replacement Details}
\label{sec:appendix:anonymization}
We follow the following rules when replacing symbols:
\begin{itemize}[noitemsep]
\item Only the proof symbols are being replaced as there is no need to replace both statement and proof symbols. 
\item We replace symbols only if they appear in both the statement and the proof. 
\item If the symbol $a$, for example, is mapped to $b$, we will map $A$ to $B$, and vice versa.
\item We do not replace the double-struck letters, e.g., $\mathbb{R}$, since they usually represent fields and constant. We do not replace standard constant symbols such as $\pi$.
\ignore{\item We observe that some variables carry meaning. For example, the variable $dim$ can be interpreted as $d*i*m$ or as one variable named $dim$. To avoid this ambiguity, we find all the variables' names that are longer than two English letters and use an English spellchecker\footnote{\url{https://pyenchant.github.io/pyenchant/}} to detect which of those are English words. Suppose a multiple-letter variable is detected as an English word. In that case, we will add it to the denylist to ensure we do not treat it as multiple single-letter variables' names.}
\end{itemize}

\section{Prediction Visualization Examples}

\label{sec:appendix}

\definecolor{myred}{RGB}{255, 160, 195}
\definecolor{myblue}{RGB}{86, 201, 255}

\DeclareRobustCommand{\r}[1]{{\sethlcolor{myred}{\hl{#1}}}}
\DeclareRobustCommand{\b}[1]{{\sethlcolor{myblue}\hl{#1}}}
\DeclareRobustCommand{\o}[1]{{\sethlcolor{orange}\hl{#1}}}
\DeclareRobustCommand{\om}[1]{\colorbox{orange}{$\displaystyle #1$}}
\DeclareRobustCommand{\rm}[1]{\colorbox{myred}{$\displaystyle #1$}}
\DeclareRobustCommand{\bm}[1]{\colorbox{myblue}{$\displaystyle #1$}}


We provide a LIME visualization of several mathematical statements in Figure~\ref{fig:qual}.

\end{document}